\documentclass{article}
\usepackage{spconf,amsmath,graphicx}
\usepackage{adjustbox}
\usepackage{xcolor}
\usepackage{subcaption}
\usepackage{hyperref}

\title{Robust Unsupervised Multi-Object Tracking in Noisy Environments}
\name{C.-H. Huck Yang$^{1\dag}$ \thanks{$^{\dag}$A part of this work was done during Huck's visit to Hitachi Central Research Lab, Tokyo, Japan. Datasets are available at \url{https://github.com/huckiyang/MOT-Kuzushiji-Fashion-Video}.}, Mohit Chhabra$^{2}$, Y.-C. Liu$^{1}$, Quan Kong$^{2}$, Tomoaki Yoshinaga$^{2}$, Tomokazu Murakami$^{2}$ }
\address{$^{1}$Georgia Institute of Technology, Atlanta, GA, USA \\ $^{2}$Lumada Data Science Lab. Hitachi, Ltd., Tokyo, Japan}

\begin{document}
\maketitle
\begin{abstract}
Physical processes, camera movement, and unpredictable environmental conditions like the presence of dust can induce noise and artifacts in video feeds. We observe that popular unsupervised MOT methods are dependent on noise-free inputs. We show that the addition of a small amount of artificial random noise causes a sharp degradation in model performance on benchmark metrics.  We resolve this problem by introducing a robust unsupervised multi-object tracking (MOT) model: AttU-Net. The proposed single-head attention model helps limit the negative impact of noise by learning visual representations at different segment scales. AttU-Net shows better unsupervised MOT tracking performance over variational inference-based state-of-the-art baselines. We evaluate our method in the MNIST-MOT and the Atari game video benchmark. We also provide two extended video datasets: ``Kuzushiji-MNIST MOT'' which consists of moving Japanese characters and ``Fashion-MNIST MOT'' to validate the effectiveness of the MOT models. 
\end{abstract}
\section{Introduction}
\label{sec:intro}
Multi-object tracking (MOT) is a challenging video processing task, where the aim is to locate and track one or more visual pattern(s) from an input video. MOT has many essential real-world applications, such as multimedia content analysis~\cite{johansson2007specification} and industrial automation~\cite{grigorescu2020survey}. However, the time and pecuniary cost of labeling large-scale video data constrains the effectiveness of current methods. Unsupervised Multi-Object Tracking (UMOT) is an appealing approach to further improve performance without the requirement of labeled training data. 

\begin{figure} %
\begin{center}
   \includegraphics[width=0.80\linewidth]{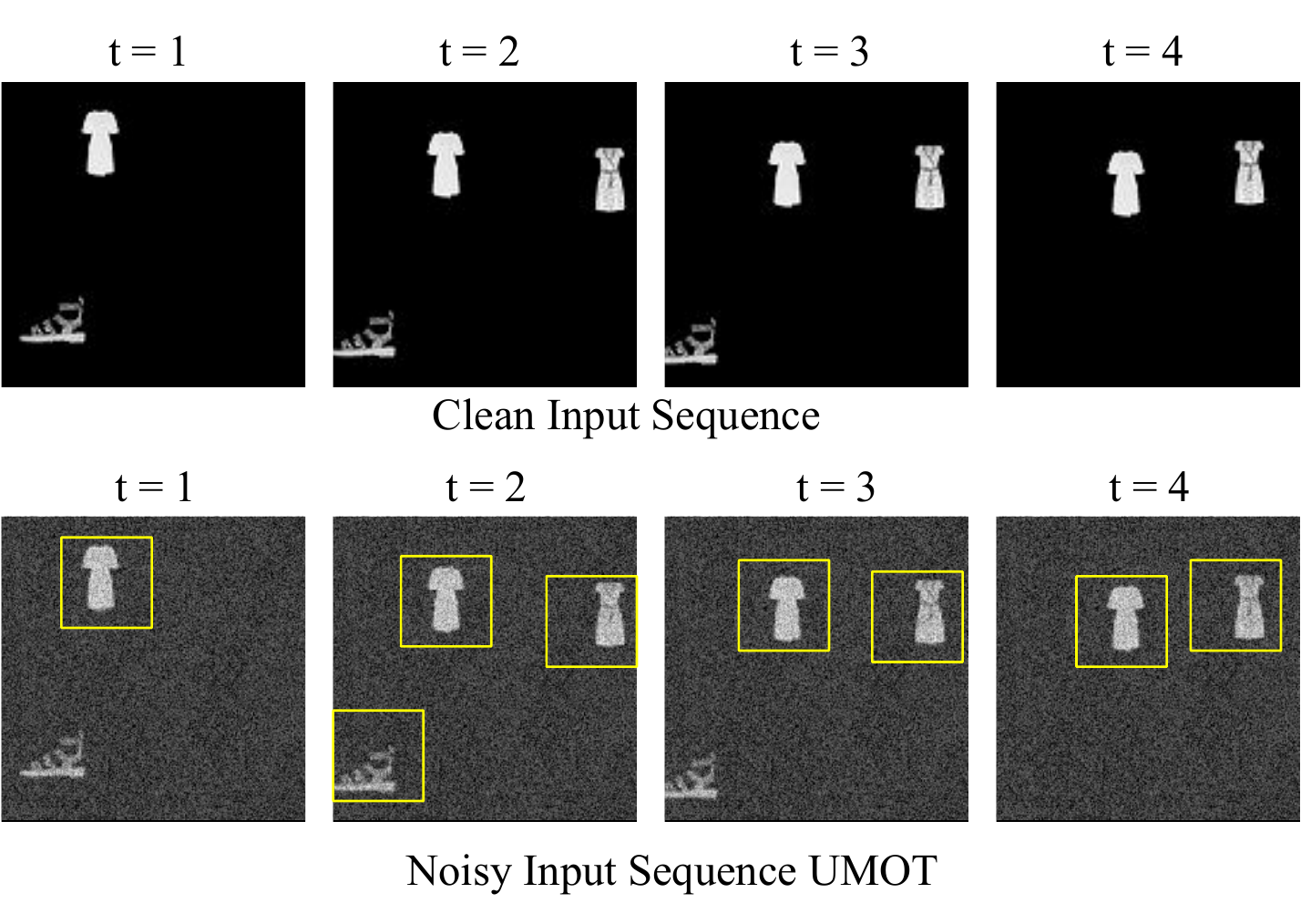}
\end{center}
\vspace{-0.7cm}
\caption{Unsupervised Multi-Object Tracking (UMOT) results on Fashion-MNIST MOT dataset based on Fashion-MNIST \cite{xiao2017fashion}. Experimental setup is motivated by ~\cite{he2019tracking, crawford2019spatially}.} 
\label{fig:figure1}
\vspace{-0.5cm}
\end{figure}

Real-world applications often require data processing in the presence of \textbf{environmental noise} as shown in Figure~\ref{fig:figure1} which is different from UMOT evaluations of video data in carefully controlled laboratory settings. The robustness of UMOT performance against environmental noise also remains relatively unexplored. In this work, we first evaluate the robustness of the state-of-the-art UMOT model against artificial noise. We propose a multi-scale tracker based on attention U-Net to improve model generalization and to avoid over-fitting on background pixels. We also show ablation studies on different visual and optimization setups in the MNIST-MOT benchmark and two extended video datasets. Our contributions include:
\begin{itemize}
    \item Investigation of the effect of noise on UMOT performance.
    \item A new DNN video tracking backbone to leverage upon attention U-Net for robust UMOT.
    \item Evaluation of the proposed method on MNIST-MOT ~\cite{he2019tracking}, Atari gaming video~\cite{crawford2020exploiting}, and two newly created UMOT datasets with complex patterns: (1) Kuzushiji-MNIST MOT derived from \cite{clanuwat2018deep} and (2) Fashion-MNIST MOT derived from \cite{xiao2017fashion}.
\end{itemize}
\section{Related Work}
\label{sec:ref}
\subsection{Unsupervised Multi-Object Tracking}
Unsupervised object tracking often incorporates visual reconstruction and geometric rendering to utilize spatial information for predicting bounding boxes. Tracking by animation~\cite{he2019tracking} (TBA) showed the first competitive UMOT results on both the MNIST-MOT and DukeMTMC~\cite{gou2017dukemtmc4reid} dataset. It used a recurrent neural network (RNN) to learn the representations of object motion and a convolutional neural network (CNN) based autoencoder to reconstruct frames and visual patterns from geometrical tracker arrays. Spatially invariant attend-infer-repeat (SPAIR) models~\cite{crawford2019spatially} used CNN feature extractors and a local spatial object specification scheme to conduct UMOT video processing with variational inference. Building upon the SPAIR method, spatially invariant label-free object tracking (SILOT) ~\cite{crawford2020exploiting} incorporated VAE based architecture with competitive MOT accuracy on MNIST-MOT and Atari video games. 
However, previous methods and benchmark models do not provide experimental studies on the effect of background noise on tracker performance. This motivates the evaluation setup under noisy conditions.
\subsection{Self attention for robustness}
Image denoising~\cite{buades2005review} with the objective of image enhancement has been studied extensively in the literature (e.g., image dehazing~\cite{yang2020net}). Image de-noising approaches have also been investigated in the context of improving representation learning. U-Net~\cite{ronneberger2015u} was initially introduced for the task of biomedical image segmentation with few labeled images. 
U-Net based multi-scale learning architectures, such as Wave-U-Net~\cite{yang2019wavelet} and attention U-Nets~\cite{oktay2018attention, yang2020wavelet}, have attained competitive results against noisy inputs by learning across multiple scales over input segments. Siam-U-Net~\cite{dunnhofer2020siam} has shown that U-Net based encoders perform well on visual tracking tasks. In this work, we build upon U-Net based backbone for UMOT. The motivation is to enhance model generalization with multi-scale de-noising. \begin{figure*}[ht!]
\begin{center}
   \includegraphics[width=1.00\linewidth]{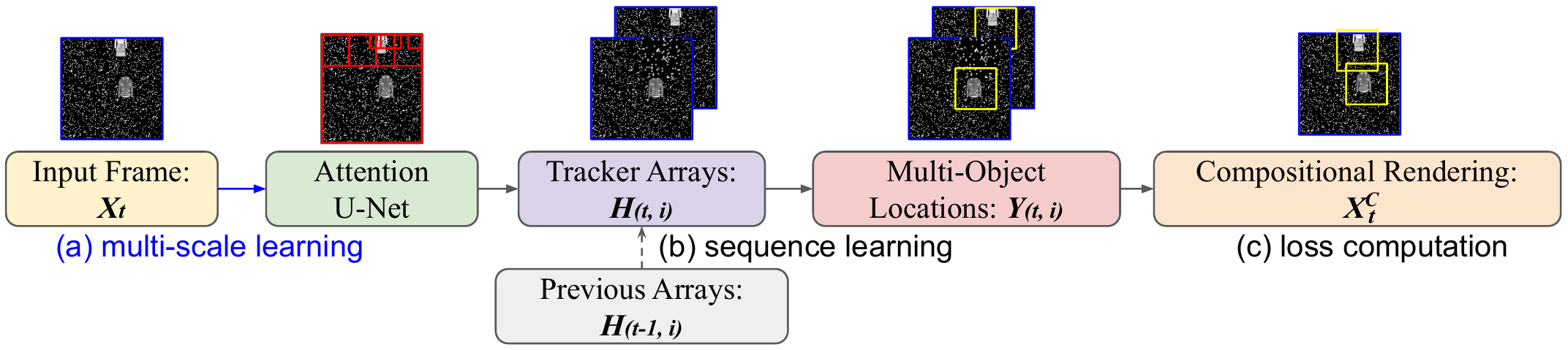}
\end{center}
\vspace{-0.7cm}
   \caption{Overview of the U-Net enhanced animation tracking frame-work: (a) multi-scale learning; (b) sequence learning; (c) loss computation. This enhanced scale-free learning method is motivated by the tracking-by-animation benchmark~\cite{he2019tracking}. } 
\label{fig:figure2}
\vspace{-0.5cm}
\end{figure*}

\section{Enhanced Unsupervised Multi-Object Tracking}
\subsection{Noisy Background Setup}
The loss function $l_t$ at time $t$ is defined similarly to the loss function in TBA (refer to Eq. (9) in ~\cite{he2019tracking}):  
\begin{equation}
l_{t}=\operatorname{MSE}\left(\boldsymbol{X}_{t}, \boldsymbol{X}^{C}_{t}\right)+\lambda \cdot \frac{1}{I} \sum_{i=1}^{4} (s_{t, i}^{x}, s_{t, i}^{y}),
\label{eq:1}
\end{equation}
where the first term is the mean squared error (MSE) between $\boldsymbol{X}_t$, a ground truth frame and $\boldsymbol{X}^{C}_t$, reconstructed frame generated by DNN. The second term is tightness constraint on the bounding box size and is computed by $\lambda$, a scaling coefficient; $I$, the number of trackers and $s_{t, i}^{x}, s_{t, i}^{y}$ object poses. 

To simulate noisy conditions, we consider random noise $\delta_{t} \sim \mathcal{N}(0,1)$ sampled from Gaussian distribution, which has been used previously in robust video learning studies~\cite{yang2020enhanced, lin2017tactics, grigorescu2020survey}. The total training frames of a video input in Eq. (\ref{eq:1}) are modified to $\sum^{T}_{t=1}\boldsymbol{X}{'}_t = \sum^{T}_{t=1}(\boldsymbol{X}_t + \beta \times \delta_{t})$ as a \textbf{noisy setup in testing} for total time step $t\in\{1,2,...,T\}$, where $\beta \in \{0\%,10\%, 20\%, 30\% \}$ scales the noise magnitude to a percentage of maximum intensity (refer to \cite{yang2020enhanced,lin2017tactics}.
\subsection{Attention U-Net Feature Encoder}
\label{sec:attu}
    We design a spatial feature encoder consisting of transformer-based single-head attention~\cite{yang2020characterizing, vaswani2017attention} module. ResNet$_{18}$~\cite{he2016deep} encoder extracts the features map ($m_t$) from inputs $\boldsymbol{X_{t}}$. The feature map $m_t$ is also used to extract key $k_t$, value ($v_t$) and query ($q_t$) tensors. The attention map ($A_t$) is computed by scaling the value ($v_t$) tensor by the inner product of the key ($k_t$) and query ($q_t$) tensors along the spatial dimensions (width and height) of the feature map. Here, the $\operatorname{view}$ operation is used to reshape the tensors.
\begin{align}
    m_t =&~f_{ResNet,\theta_1}(\boldsymbol{X_{t}}); \quad q_t =~\operatorname{view}\left(f_{q,\theta_2}(m_t)\right);\\
     \quad k_t =&~\operatorname{view}\left(f_{k,\theta_3}( m_t)\right);v_t =~\operatorname{view}\left(f_{v,\theta_4}( m_t)\right); \\
     \quad A_t =&~\operatorname{view}\left(\frac{q_t k_{t}^{T}}{\sqrt{d_{k}}}\right) v_t,
\end{align}
where $f_{ResNet}$, $f_q$, $f_k$, $f_v$ are individual DNNs with hyper-parameters ($\theta_1$ to $\theta_4$) updated by end-to-end gradient descent training. 
As shown in Figure \ref{fig:figure2} (a), a U-Net~\cite{ronneberger2015u} encoder is used to learn multi-scale features from the attention map:
$C_t = \text{U-Net}(A_t; \theta_5)$.
 The attention U-Net comprises of channel sizes $\{32,64,128,64,32\}$. Each convolutional layer is followed by a rectified linear unit and a $2\times2$ max-pooling operation with stride $2$ for down-sampling.
\subsection{Tracker Arrays and Rendering}
After obtaining the feature representation $C_{t}$, we use tracker arrays to extract the individual location of the object(s) from a frame. As shown in Figure \ref{fig:figure2} (b), we use a latent state $\boldsymbol{H}_{(t,i)}$ feature extractor for $i^{th}$ tracker to convert spatial information by neural sequence modeling:

\begin{equation}
   \boldsymbol{H}_{(t,i)}=f_{seq,\theta_6}\left(\boldsymbol{H}_{t-1}, C_{t} \right), 
\end{equation}

where $f_{seq,\theta_6}$ consists of bidirectional GRU~\cite{chung2014empirical} networks. To obtain the location of objects in an unsupervised manner, we incentivize a jointly parameterized tracker network ($f_\text{joint}$) to predict a single object location ($\boldsymbol{Y}_{t,i}$).
\begin{equation}
\boldsymbol{Y}_{t,i}=f_\text{joint}\left(\boldsymbol{h}_{t, i} ; \boldsymbol{\theta}^{\text {out }}\right)
\end{equation}
where the output $\boldsymbol{Y}_{t,i}$ is a mid-level representation of objects on $2 \mathrm{D}$ image planes, including: (a) probability of having tracked an object; (b) object pose ($s_{t, i}^{x}, s_{t, i}^{y}$ in Eq.~(\ref{eq:1})); (c) shape (with channel size); (d) object appearance. These mid-level output locations are used to validate model performance by MOT evaluation. We follow the layer-wise composition setup from TBA~\cite{he2019tracking} and use Spatial Transformer Network (STN)~\cite{jaderberg2015spatial} to scale  $\boldsymbol{Y}_{t,i}$ to multiple layer foreground ($L^{k}_{t}$) and layer masks ($L$-$m^{k}_{t}$) with layer number $k=3$. As shown in Figure \ref{fig:figure2} (c), we contain the DNN reconstructed frame $\boldsymbol{X}^{C}_t$ by using layer-wise geometrical composition: 
\begin{align}
\boldsymbol{X}_{t}^{(k)} =&\left(\mathbf{1}-L\text{-}m^{k}_{t}\right) \odot \boldsymbol{X}_{t}^{(k-1)}+L_{t}^{k};\\
\boldsymbol{X}_{t}^{C} =& \sum_{k=1}^{3}\boldsymbol{X}_{t}^{(k)},
\end{align}
$\boldsymbol{X}_{t}^{C}$ is used for computing reconstruction loss Eq.~(\ref{eq:1}).
\section{Experiments}
\begin{figure} %
\begin{center}
   \includegraphics[width=1.00\linewidth]{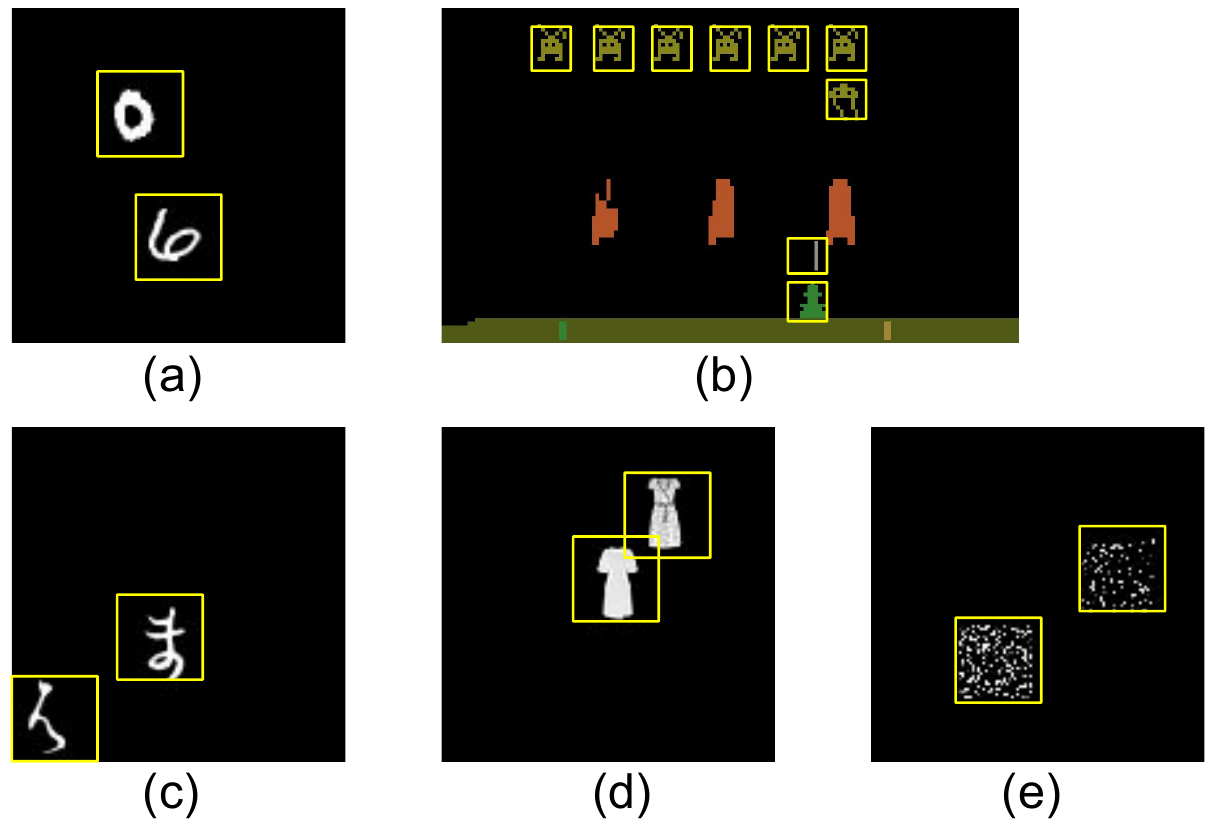}
\end{center}
\vspace{-0.7cm}
   \caption{Four training datasets used in our experiments: (a) MNIST-MOT \cite{he2019tracking, crawford2019spatially}, (b) Atari gaming video~\cite{crawford2020exploiting}, (c) Kuzushiji-MNIST MOT, and (d) Fashion-MNIST MOT. A test dataset with scrambled objects (e) is also provided to evaluate generalization ability.} 
\label{fig:figure3}
\vspace{-0.5cm}
\end{figure}

\subsection{Datasets and Baseline Setup}
\label{sec:4:1}
As shown in Figure ~\ref{fig:figure3}, we select two baseline video datasets from the previous UMOT studies: (a) MNIST-MOT ~\cite{he2019tracking, crawford2019spatially} containing $2M$ training frames and $25k$ validation frames, where each frame is of size $128\times128\times1$ and (b) Atari gaming video on Space-Invader (Atari-SI)~\cite{crawford2020exploiting, crawford2019spatially} containing $128k$ training frames and $1k$ validation frames, where each frame is converted to gray-scale with a input size of $210\times160\times1$ for testing.  

\textbf{MOT Video Datasets:} To study the effects of tracking on patterns other than MNIST characters, we provide two new MOT video datasets with more intricate patterns (c) Japanese cursive characters (denoted as \textbf{Kuzushiji} from ~\cite{clanuwat2018deep}) and (d) Fashion-MNIST MOT (denoted as \textbf{Fashion} derived from ~\cite{xiao2017fashion}). These new datasets have been chosen to study complex visual patterns and could be useful for different UMOT applications (e.g., pedestrian tracking).

\textbf{Object-Scrambling Test:} To verify whether the unsupervised model indeed learns semantically meaningful patterns, we provide an \textbf{object-scrambling} video test-set  containing $1k$ frames as shown in Figure~\ref{fig:figure3} (e). The test-set was constructed from the MNIST-MOT video dataset by randomly permuting the pixels in object bounding boxes.  Ideally, a robust UMOT model~\cite{crawford2019spatially, grigorescu2020survey} should learn semantically consistent discriminative representations for predicting object location. A model which learns such disciminative patterns on the MNIST-MOT dataset is expected to have degraded performance on the object scramble test-set.

\textbf{Baseline Models:} We select two major benchmark UMOT models as reproducible studies as shown in Table~\ref{tab:1}. 
\begin{itemize}
    \item Tracking by animation (TBA)~\cite{he2019tracking}: a model combines with RNNs for sequence modeling with attention features and uses CNN feature extraction for representation learning on frame rendering. 
    \item Spatially invariant label-free object tracking (SILOT)~\cite{crawford2020exploiting}: a model built upon VAE for frame-wise reconstruction and using YOLO-based CNN feature extraction for representation learning. 
\end{itemize}
\textbf{U-Net based UMOT backbones:} We design two different type of U-Net based feature encoder backbones: residual U-Net network with a ResNet$_{50}$ encoder (denoted as \textbf{ResU-Net}) and propose self-attention U-Net (denoted as \textbf{AttU-Net}) as discussed in section \ref{sec:attu}. 
The tested UMOT models have a comparable number of model parameters $\sim12M$.   
\begin{table*}[ht!]
\centering
\caption{Detailed MOT metrics in the clean MNIST-MOT dataset~\cite{he2019tracking, crawford2020exploiting} with UMOT models as a reproducible baseline.}
\label{tab:1}
\begin{adjustbox}{width=0.95\textwidth}
\begin{tabular}{|l|l|l|l|l|l|l|l|l|l|l|l|l|l|l|l|}
\hline
Model & IDF1 & IDP & IDR & Rcll & Prcn & GT & MT & PT & ML & FP & FN & IDs & FM & MOTA & MOTP \\ \hline \hline
TBA & 99.4\% & 99.3\% & 99.5\% & 99.8\% & 99.6\% & 99 & 99 & 2 & 0 & 9 & 4 & 2 & 1 & 99.3\% & 88.4\% \\ \hline
SILOT & 98.9\% & 98.2\% & 99.5\% & 99.9\% & 98.5\% & 100 & 100 & 0 & 0 & 34 & 3 & 7 & 1 & 98.0\% & 85.9\% \\ \hline
ResU-Net & 99.5\% & 99.5\% & 99.8\% & 100.0\% & 99.7\% & 99 & 99 & 0 & 4 & 3 & 2 & 1 & 0 & 99.4\% & 89.2\% \\ \hline
AttU-Net & 99.9\% & 99.9\% & 100.0\% & 100.0\% & 99.9\% & 99 & 99 & 0 & 0 & 3 & 1 & 1 & 0 & 99.8\% & 89.6\% \\ \hline
\end{tabular}
\end{adjustbox}
\end{table*}
\subsection{UMOT Performance in Different Contexts}
\begin{figure}[ht!]
        \centering
        \begin{subfigure}[b]{0.230\textwidth}
            \centering
            \includegraphics[width=\textwidth]{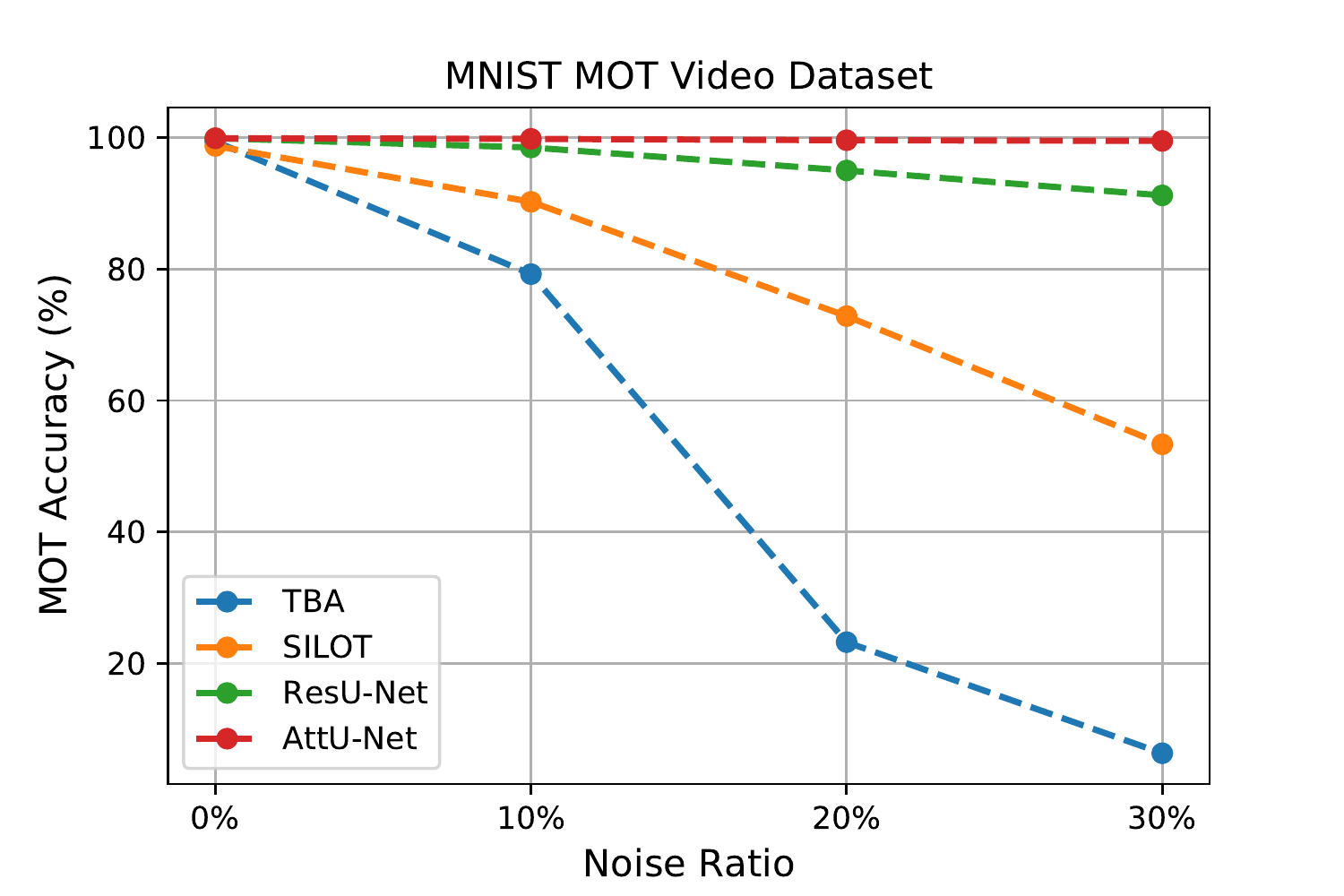}
            \caption[Network2]%
            {{\small MNIST}}    
            \label{fig:mnist}
        \end{subfigure}
        \hfill
        \begin{subfigure}[b]{0.230\textwidth}  
            \centering 
            \includegraphics[width=\textwidth]{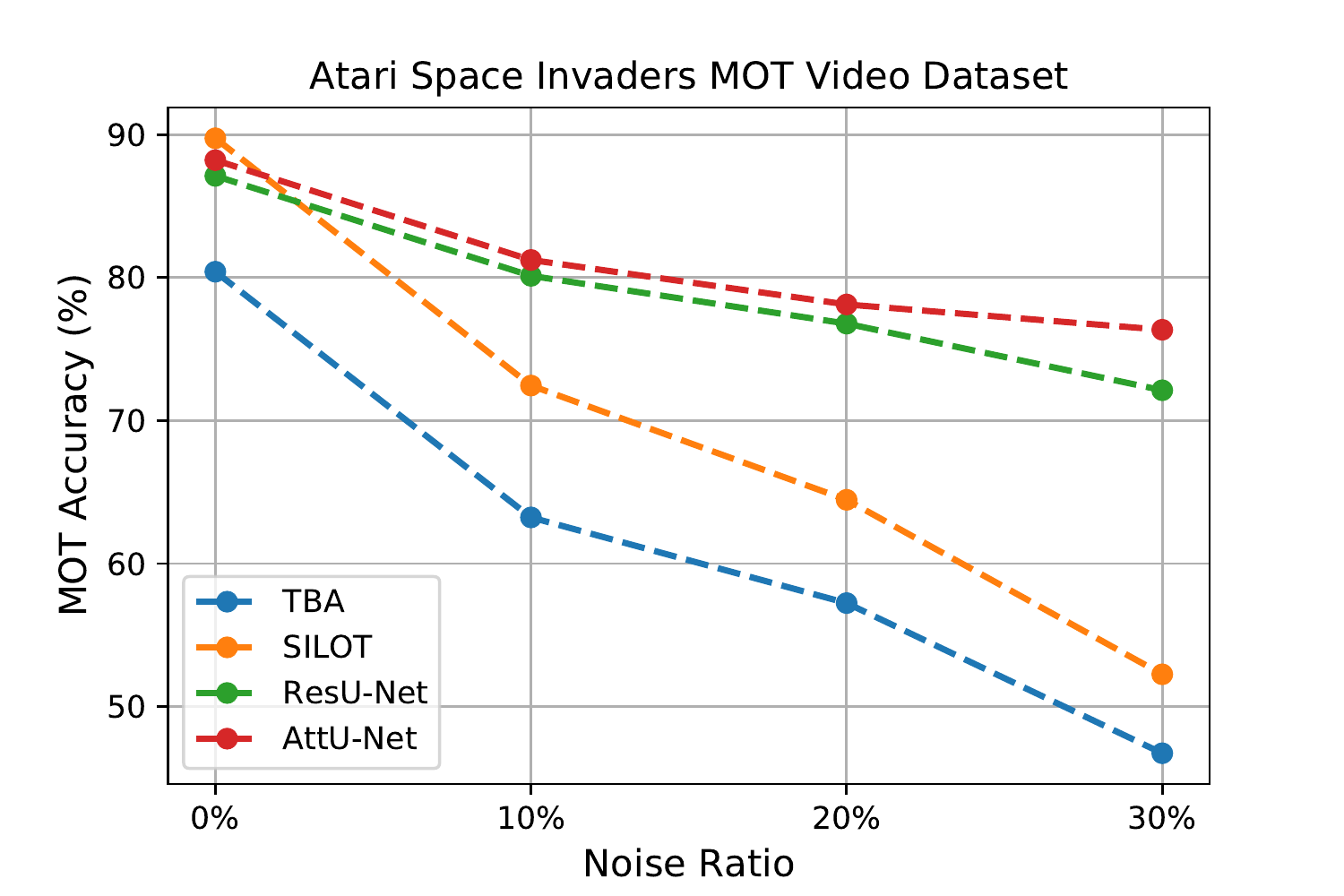}
            \caption[]%
            {{\small Atari-SI}}    
            \label{fig:atari}
        \end{subfigure}
        \vskip\baselineskip
        \begin{subfigure}[b]{0.230\textwidth}   
            \centering 
            \includegraphics[width=\textwidth]{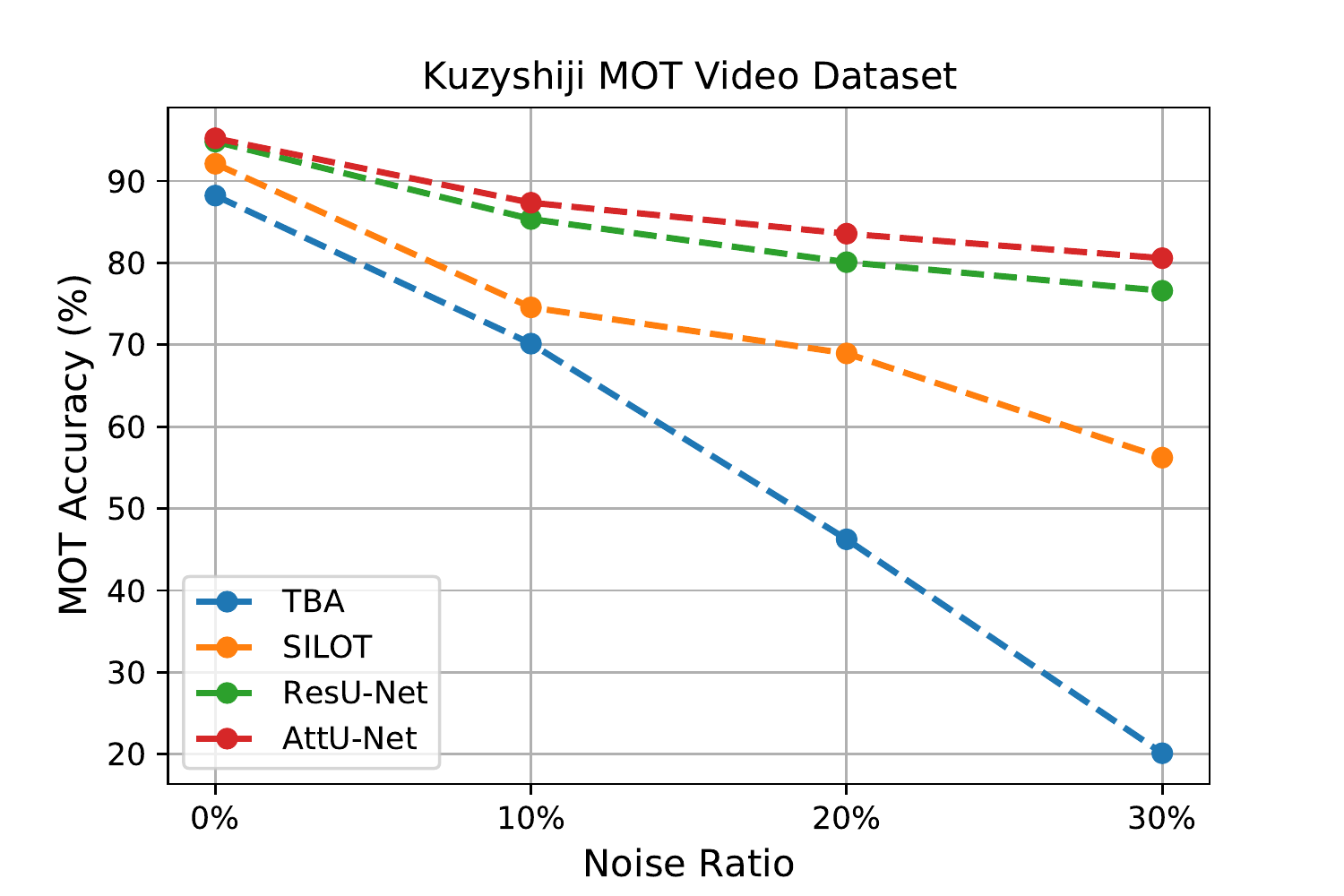}
            \caption[]%
            {{\small Kuzushiji}}    
            \label{fig:kuzushiji}
        \end{subfigure}
        \hfill
        \begin{subfigure}[b]{0.230\textwidth}   
            \centering 
            \includegraphics[width=\textwidth]{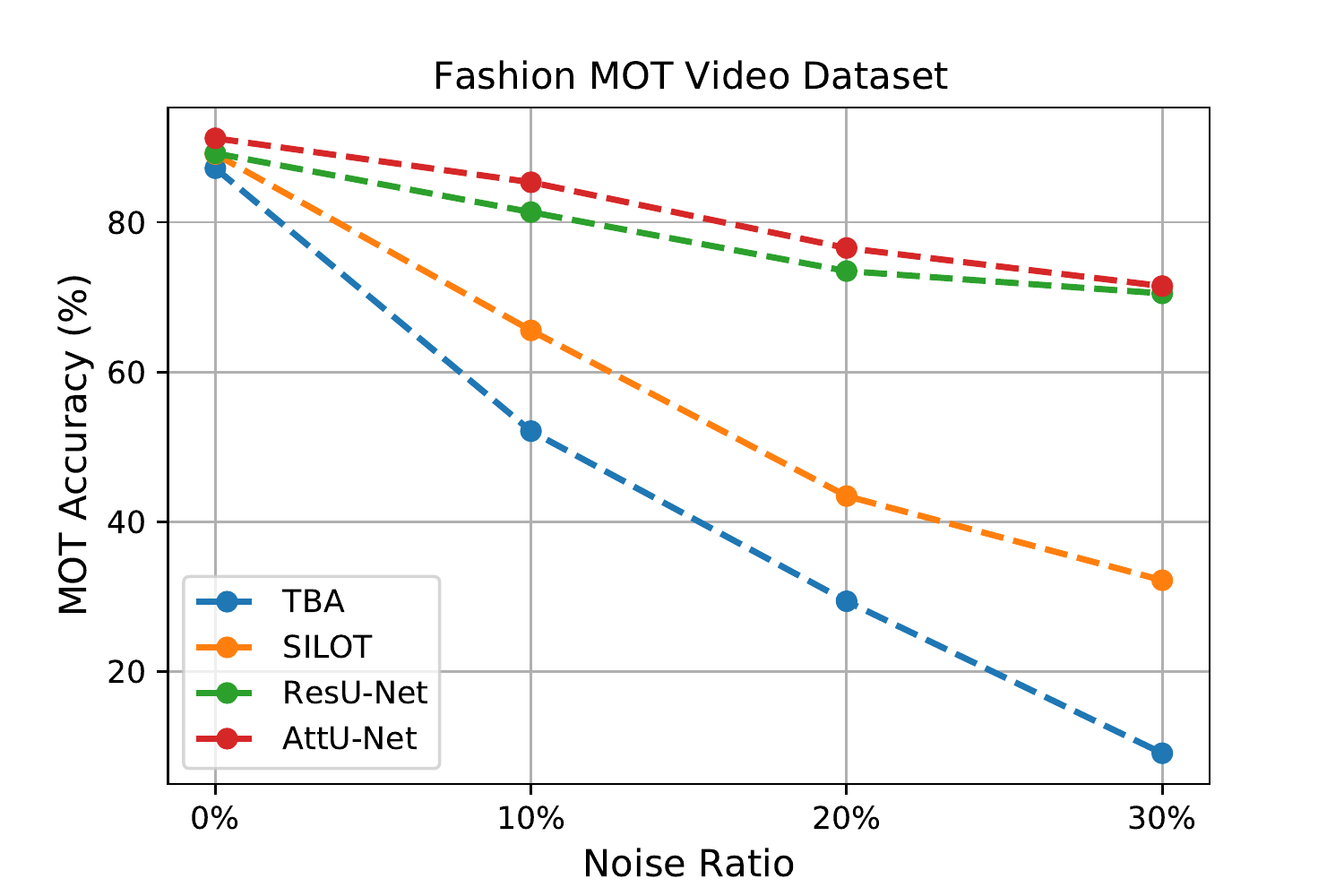}
            \caption[]%
            {{\small Fashion}}    
            \label{fig:fashion}
        \end{subfigure}
        \caption[ The average and standard deviation of critical parameters ]
        {\small MOTA Performance under different noisy level testing in: (a) MNIST-MOT ~\cite{crawford2019spatially, he2019tracking}, (b) Atari-Space Invader~\cite{crawford2020exploiting} (SI) video, (c) Kuzushiji-MNIST MOT, and (d) Fashion-MNIST MOT tracking datasets.} 
        \label{fig:noise}
    \end{figure}

 As shown in Figure~\ref{fig:noise}, we evaluated the MOT accuracy with noise level ranging from $0\%$ (without noise) to $30\%$. All the models were trained by using noise free frames as shown in Figure~\ref{fig:figure3} (a) to (d). In the noise free setting ($0\%$ noise), MNIST-MOT derived from (\ref{fig:mnist}) and Kuzushiji-MOT derived from (\ref{fig:kuzushiji}), the  evaluated models attained MOT accuracy ranging above $95$; Atari-SI (\ref{fig:atari}) and Fashion-MNIST MOT derived from (\ref{fig:fashion}) were found to be more challenging and the MOT accuracy was below $90\%$ for tested models. When the the noise magnitude was increased, TBA model showed a sharp decrease in performance over all tested datasets. MOT accuracy for TBA model fell below $\sim 20\%$ in the MNIST-MOT dataset as noise was increased to $30\%$. 

Although SILOT models are more robust than TBA under the presence of small amount of noise. Their performance suffers major degradation when the noise ratio is increased. It decreases to $52.12\%$ in Atari-SI when the noise level is $20\%$. For U-Net based backbones, both ResU-Net and AttU-Net show robustness in the MNIST-MOT dataset even with the presence $30\%$ noise. For all datasets, AttU-Net based UMOTs show best performance with substantial absolute improvements ($\sim3$-$5\%$) over the ResU-Net with the exception of Fashion-MNIST MOT where the improvement is ($1.01\%$) as (\ref{fig:fashion}).
In short, the performance of existing benchmark UMOTs, TBA and SILOT, are sensitive to noise whereas U-Net based UMOTs perform competitively. 
\subsection{Evaluation of UMOT Model Generalization}
 To study generalization of different UMOT models with noisy inputs, we evaluated pre-trained UMOT models on MNIST-MOT test set in two different configurations. 
 \textbf{(1) Testing with scrambled objects:} As in Table~\ref{tab:src}, U-Net backbones show major degradation when context-wise representation has been corrupted by the scrambled object settings ($S$). As discussed in Section \ref{sec:4:1} U Net backbones generalize better than baseline models.
\textbf{(2) Training with augmented noise: } We also evaluated as to how models would benefit from training with noisy Fashion-MNIST MOT video dataset. As shown in Table~\ref{tab:noise}, both U-Net based UMOT models show better performance than TBA and SILOT by a noisy training ($N$) with augmented inputs containing $30\%$ noise ratio. All UMOTs improve in terms of absolute performance but the architecture-specific (e.g., multi-scale learning by U-Net based backbones) differences still significantly impact performance metrics.  
\begin{table}[ht!]
\centering
\caption{Performance of pretrained UMOT models on MNIST-MOT when tested on scrambled MNIST objects ($S$). Higher numbers indicates lack of semantic context.}
\label{tab:src}
\begin{tabular}{|c|c|c|c|c|}
\hline
 & TBA~\cite{he2019tracking} & SILOT~\cite{crawford2020exploiting} & ResU-Net & AttU-Net \\ \hline \hline
$S$-MOTA & 91.4\% & 83.9\% & 65.1\% & \textbf{61.7\%} \\ \hline
$S$-MOTP & 90.9\% & 82.1\% & 61.4\% &\textbf{ 59.8\%} \\ \hline
\end{tabular}
\end{table}
\vspace{-4mm}
\begin{table}[ht!]
\centering
\caption{Performance of different UMOT models trained with $30\%$ noise ($N$) in Fashion-MNIST MOT dataset. Higher numbers indicate better generalisation ability. }
\label{tab:noise}
\begin{tabular}{|c|c|c|c|c|}
\hline
 & TBA~\cite{he2019tracking} & SILOT~\cite{crawford2020exploiting} & ResU-Net & AttU-Net \\ \hline \hline
$N$-MOTA & 24.1\% & 46.7\% & 76.3\% & \textbf{81.2}\% \\ \hline
$N$-MOTP & 23.2\% & 44.8\% & 75.2\% & \textbf{80.1}\% \\ \hline 
\end{tabular}
\end{table}
\vspace{-4mm}

\section{Conclusion}
In this work, we studied UMOT performance with noisy inputs and show that pre-existing methods suffer a substantial decrease in MOT accuracy when the noise ratio is increased above $30\%$. AttU-Net based approach increases the robustness to noise on all evaluated data sets.  We also show the ability of AttU-Net to learn semantically relevant features by testing it on the test set of scrambled instances.

\clearpage
\bibliographystyle{IEEEbib}
\bibliography{strings,refs}

\end{document}